\documentclass[10pt, conference, compsocconf]{IEEEtran}
\ifCLASSINFOpdf
\else
\fi
\usepackage{graphicx}
\usepackage{subcaption}

\begin{document}
%
\title{Survey of Face Detection on Low-quality Images}


\author{\IEEEauthorblockN{Yuqian Zhou, Ding Liu, Thomas Huang}
\IEEEauthorblockA{Beckmann Institute,
University of Illinois at Urbana-Champaign, USA\\
\{yuqian2, dingliu2\}@illinois.edu \hspace{5mm} huang@ifp.uiuc.edu}
}


%


\maketitle

\begin{abstract}
Face detection is a well-explored problem. Many challenges on face detectors like extreme pose, illumination, low resolution and small scales are studied in the previous work. However, previous proposed models are mostly trained and tested on good-quality images which are not always the case for practical applications like surveillance systems. In this paper, we first review the current state-of-the-art face detectors and their performance on benchmark dataset FDDB, and compare the design protocols of the algorithms. Secondly, we investigate their performance degradation while testing on low-quality images with different levels of blur, noise, and contrast. Our results demonstrate that both hand-crafted and deep-learning based face detectors are not robust enough for low-quality images. It inspires researchers to produce more robust design for face detection in the wild.

\end{abstract}


%
\IEEEpeerreviewmaketitle

\section{INTRODUCTION}
Face detection has been intensively studied in the past decades because of its wide applications in face analysis. As an important processing step for face recognition, a robust detection algorithm is expected to identify faces under arbitrary image conditions. Previous work has demonstrated robustness in face conditions like extreme poses, multiple face scales, and occlusions. However, in the practical surveillance systems, the face detectors should have the capability of detecting faces in low-quality images with distortions like blur, noise and low contrast. Therefore, it is necessary to evaluate the performance of existing face detection algorithms on images with various distortions.

 \begin{figure}
      \centering
      \includegraphics[width=1\linewidth]{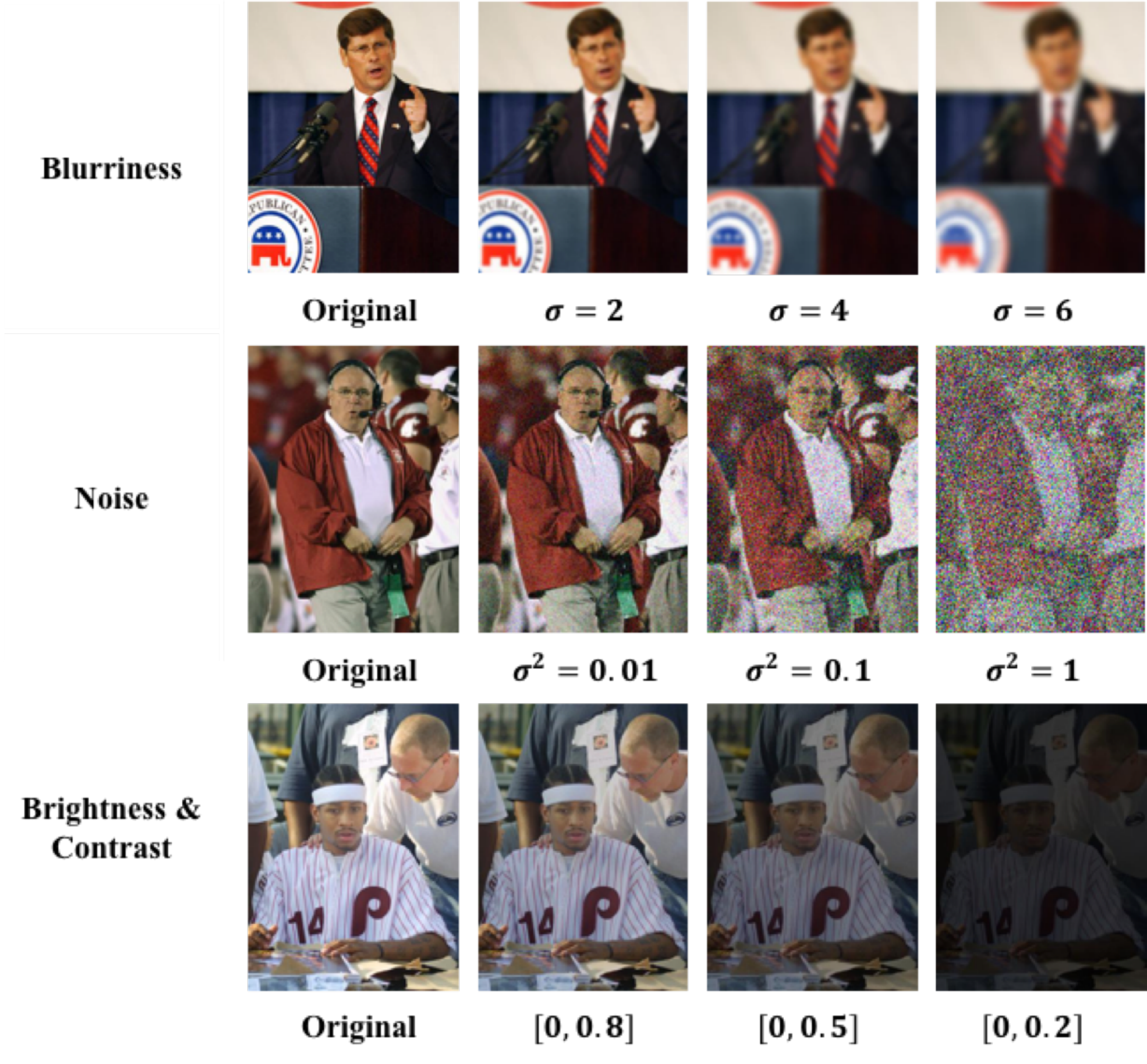}
       
      \caption{Examples of synthetic low-quality face images. For blur, we applied Gaussian blur with various standard deviations. For noise, we utilized additive Gaussian white noise. We also decrease the range of image pixel values to lower the brightness and contrast level.  }
      \label{fig:process}
      \vspace{-6mm}
   \end{figure}
   
Face detection algorithms have evolved from utilizing hand-crafted features like Haar \cite{vj_face} or SURF \cite{surfcascade} to deeply learned ones. Benefiting from large model capacity, deep learning methods generally improve the detection of large variations of faces like extreme poses and heavy occlusions by learning from large-scale data. A number of approaches based on deep Convolutional Neural Networks (CNNs) focus on handling the problem of detecting multi-scale faces, especially finding tiny faces in the images. To cope well with multi-scale problem, face detection is usually regarded as a special case of object detection with only one class. Therefore, face detection algorithms mostly follow the approaches of generic object detection and can be categorized into faster R-CNN \cite{ren2015faster}\slash R-FCN \cite{rfcn} family, and SSD \cite{liu2016ssd} family. The corresponding state-or-the-art algorithms have achieved both accurate and fast detection on multi-scale faces.

In practical applications like surveillance system, images containing faces are usually distorted in the process of acquisition, storage and transmission, causing the image quality degradation. Although saturating the performance on high-quality image benchmark like FDDB \cite{fddb}, most popular face detectors are not evaluated on low-quality images with distortions like blur or noise. It is shown that deep object recognition networks trained with high-quality samples are not reliable enough when being tested on low-quality images \cite{lq1}. However, the neural networks of multi-scale designs may be able to compensate the performance degradation caused by low-resolution and blur, which inspires us to study the influence of multi-scale strategies on low-quality face detection. 

In this paper, we investigate the robustness of face detection algorithms on low-quality images from FDDB with different levels of blur, noise and contrast. Specifically, we evaluate four representative face detection models: traditional hand-crafted detectors Viola-Jones Haar AdaBoost ~\cite{vj_face} and HoG-SVM \cite{dlib_face}, and deep learning based models: faster-RCNN \cite{fasterrcnnface} and $S^3$FD \cite{sfd}. We illustrate the robustness level of algorithms varying from features and multi-scale designs. We hope our results can inspire researchers to propose more quality-invariant face detectors in the future.


\section{Face Detection Algorithms}
\subsection{Traditional Methods}

Traditional face detection methods \cite{facesurvey} are based on hand-crafted features, and can be categorized into three classes: cascade methods, deformable parts model (DPM) \cite{dpm} and aggregated channel features. For cascade approaches, Viola-Jones face detector \cite{vj_face} is the milestone work with AdaBoost cascade scheme using Haar-like features. After that, more features like SURF \cite{surfcascade}, HoG \cite{dlib_face}, and LBP \cite{lbp_cascade} are investigated on a similar structure of Viola-Jones detector. Other simpler features like pixel difference in NPD \cite{npd}, Joint Cascade \cite{jointcascade} and Pico \cite{pico} etc. are developed to improve the computation speed. Another class of face detection methods based on structured models  \cite{zhu2012face,fastestDPM,headhunter,facddpm} apply DPM \cite{dpm} to cope with the intra-class variance. Most recently, researchers integrated multiple hand-crafted features \cite{integrate_channel_feature} in channels and achieved a higher accuracy. The representative work includes headhunter \cite{headhunter}, ACF-multiscale \cite{acfmultiscale}, and LDCF+ \cite{lcdf} which achieved the best performance among the traditional methods. These approaches mostly is able to achieve real-time detection on CPU, but hand-crafted features lack the robustness to complicated face variance like pose, expression, occlusion and illumination. Therefore, these methods may not be adaptive to low-quality testing samples. 

\subsection{Deep Learning Methods}

Compared to the methods using hand-crafted features, deep learning based approaches could successfully capture large variances of faces when trained on large amounts of data, thus the most challenging part becomes detecting groups of tiny faces with variance. To cope well with this problem, deep learning methods are roughly categorized into three classes: cascade CNN, faster R-CNN \cite{ren2015faster} and SSD \cite{liu2016ssd} based algorithms. Some newly proposed approaches for generic object detection like YOLO \cite{yolo}, RSA \cite{rsa}, and UnitBox \cite{unitbox} are also potential base methods for face detectors. 

Cascade CNN \cite{cascadecnn} was first proposed to address the problem of high computational cost and high variances of face detection. The intuition of cascade structure is to reject simple negative samples at early stages and refine the results later. Joint Cascade CNN  \cite{jointcascadecnn} and MTCNN \cite{mtcnn} are similar work except that they applied other facial tasks to enhance the detection. Zhang et al. proposed an ICC-CNN \cite{icccnn} to reject samples in different layers within a single CNN. The advantages of these approaches is the high computation speed. However, these methods require the usage of discrete image pyramid for multi-scale proposals, and do not explicitly resolve the problems of finding crowded, tiny and blurry faces. 

Algorithms based on Faster R-CNN \cite{ren2015faster,fasterrcnnface,facercnn} or R-FCN \cite{rfcn,rfcnface} applied a scale-invariant detector, by extracting features from ROI pooling maps in the higher layer and deploying detectors on top of that. But detecting small objects is hard using Faster R-CNN since both the background and the objects will be projected to the same pixel position in the high-level feature map. To address this problem called overlapping receptive field, CMS-RCNN \cite{cmscnn} and Deep-IR \cite{deepir} integrated features from lower-level convolutional layers to train the detector.  Utilizing low-level features also results from different visual cues used by larger and smaller faces. Approaches based on faster R-CNN achieved an impressive performance, but the computation speed is relative slow \cite{objectdetectionsurvey}. 

Algorithms based on SSD \cite{liu2016ssd} trained scale-variant detectors on different layers to take advantages of the multi-scale feature maps like in SSH \cite{ssh}. However, according to the default anchor designs of SSD, it is not suitable for detecting compact small objects. To address the anchor mismatching problem and increase the recall rate of tiny faces, $S^3$FD \cite{sfd}, FaceBoxes \cite{faceboxes}, Scaleface \cite{scaleface}, and HR-ER \cite{hrer} were recently proposed by either improving the matching strategy and anchor densities or assigning layers with specific scale ranges. Among them, $S^3$FD achieved the state-of-the-art recall in FDDB \cite{fddb} dataset. 

\section{Adversarial Testing on Deep Models}
Unluckily, deep networks for image classification tasks were proved to be sensitive to adversarial examples, which were generated by adding small perturbations using gradient methods on purpose  \cite{adversarial_attack}. These adversarial examples are hardly distinguished from the original images by human. In this case, artifacts like noise, blur, illumination or occlusion usually cause detrimental effects on the deep network performance. Extensive studies have been conducted to evaluate the effect of image distortions on deep networks \cite{wang2016studying} or hand-crafted features \cite{lqshallow}.  Dodge et al. \cite{lq1} demonstrated that VGG16 \cite{vgg} exhibited the best resilience to the image distortions compared with other deep models. Liu et al. \cite{enhancelq} attempted to resolve this problem using unsupervised pre-training and data augmentation, and achieved promising results.


\section{Experimental Setup}

\begin{figure*}[t!]
    \centering
    \begin{subfigure}[t]{0.32\textwidth}
        \includegraphics[height=1.6in]{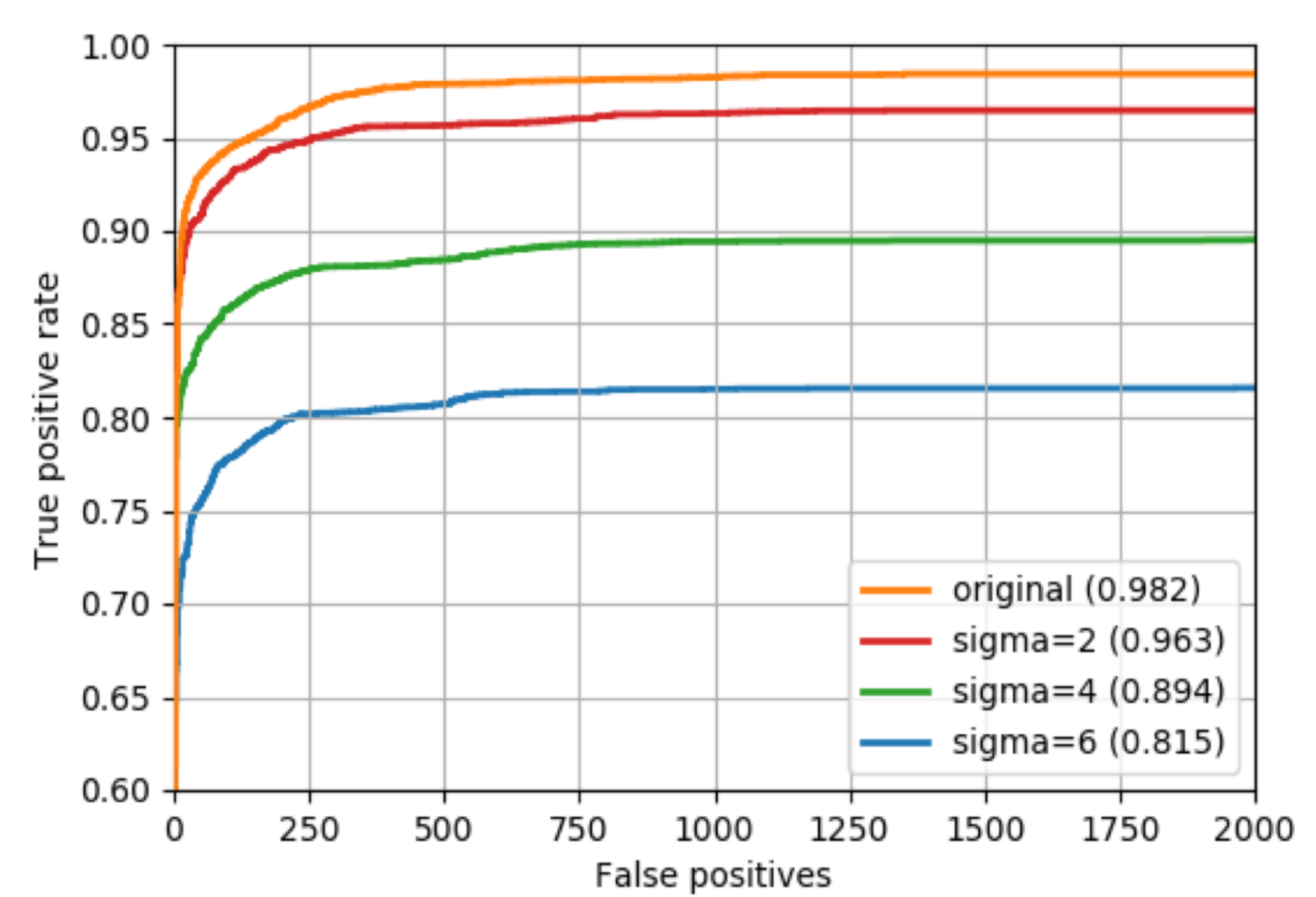}
        \caption{}
    \end{subfigure}%
    ~ 
    \begin{subfigure}[t]{0.32\textwidth}
        \includegraphics[height=1.6in]{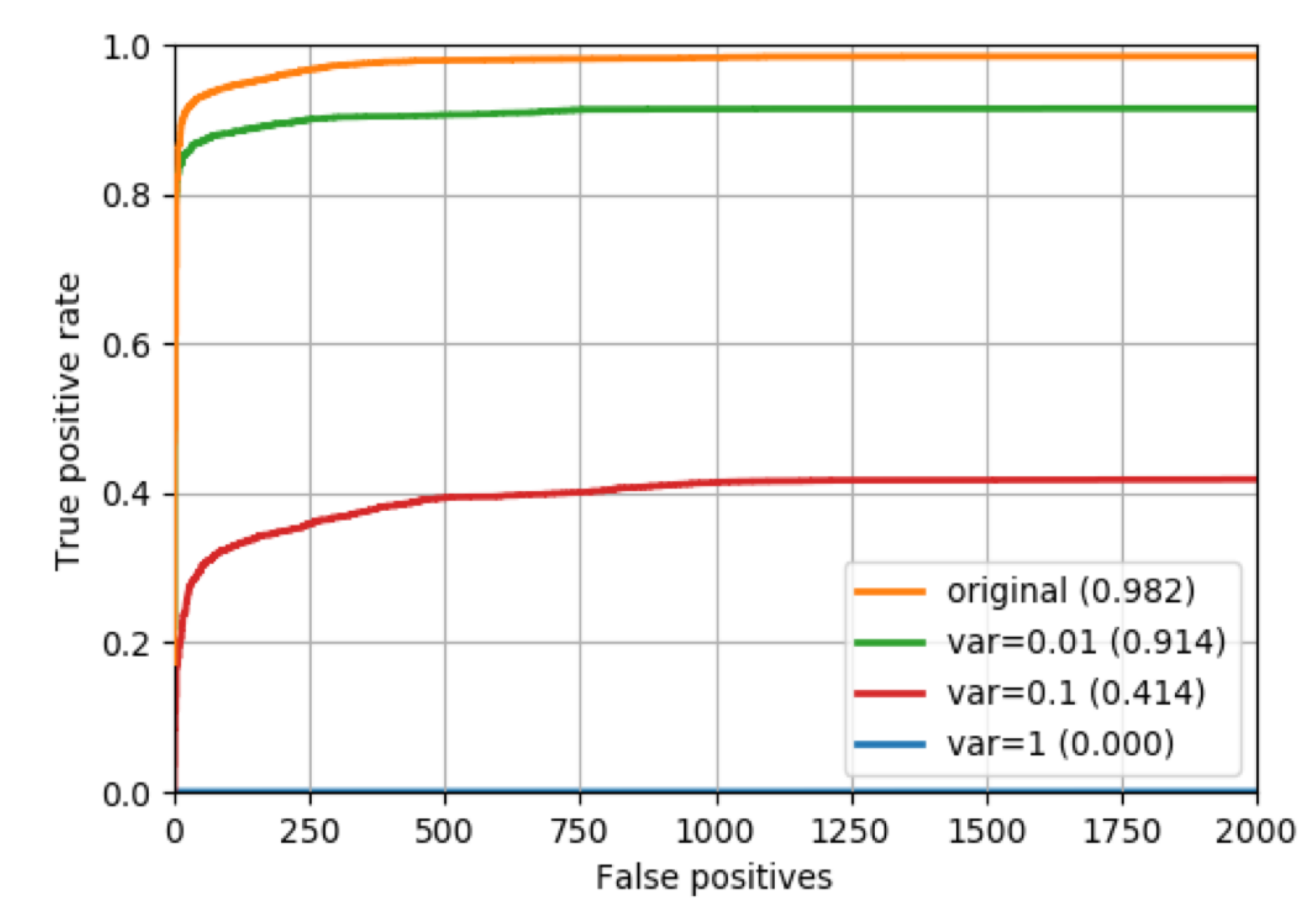}
        \caption{}
    \end{subfigure}
        ~ 
    \begin{subfigure}[t]{0.32\textwidth}
        \includegraphics[height=1.6in]{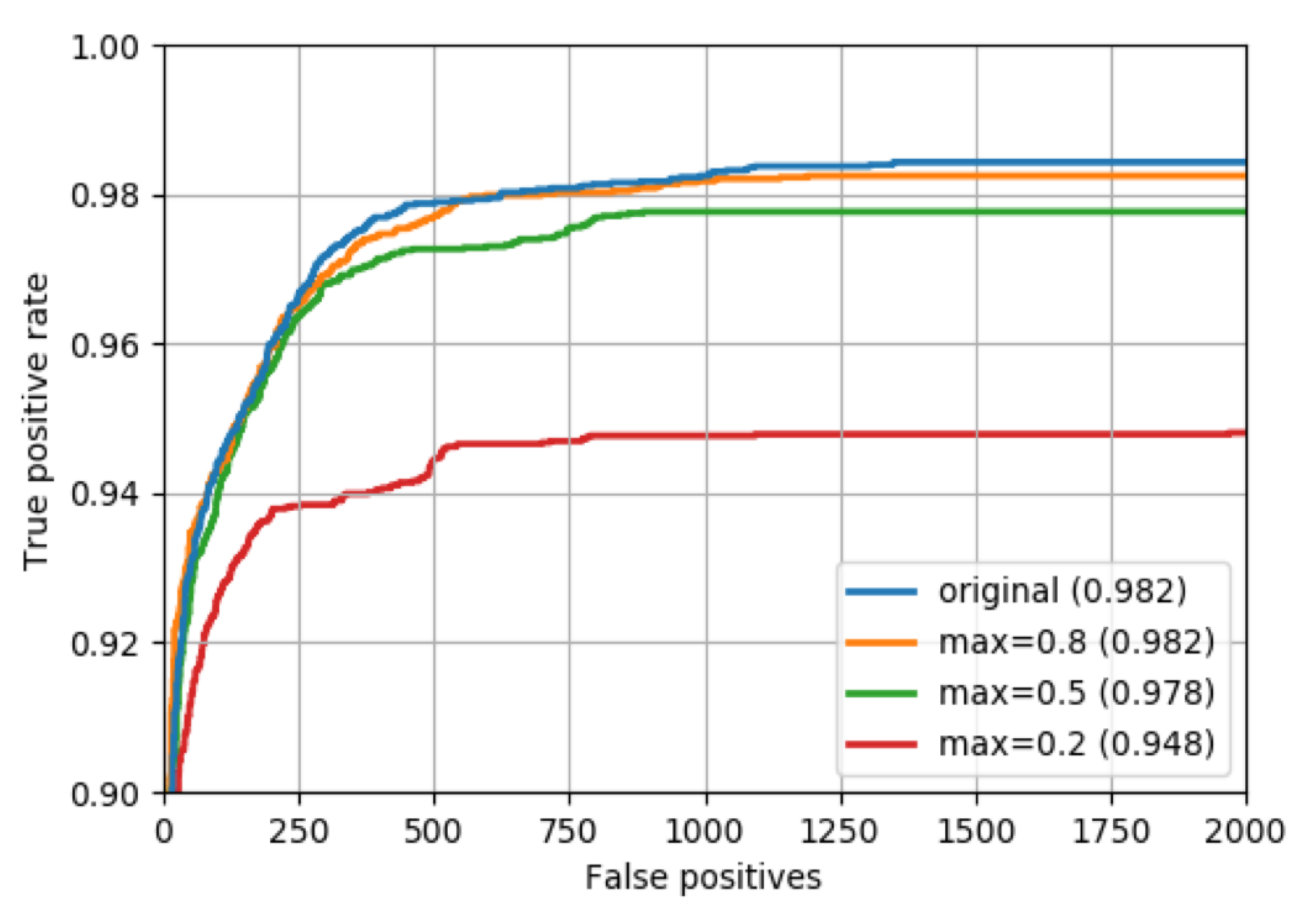}
        \caption{}
    \end{subfigure}
    \vspace{-2mm}
    \caption{Evaluation results (ROC curve) of $S^3$FD algorithm on low-quality images. Y-axis indicates the recall and X-axis represents the numbers of false positive samples. We compare the performance when (a) applying different levels of Gaussian blur, (b) adding decreasing levels of Gaussian white noise, and (c) adjusting the brightness and contrast of the whole pictures.}
    \label{sfd}
    \vspace{-5mm}
\end{figure*}

\begin{figure*}[t]
    \centering
    \hspace{-3mm}
    \begin{subfigure}[t]{0.3\textwidth}
        \includegraphics[height=1.6in]{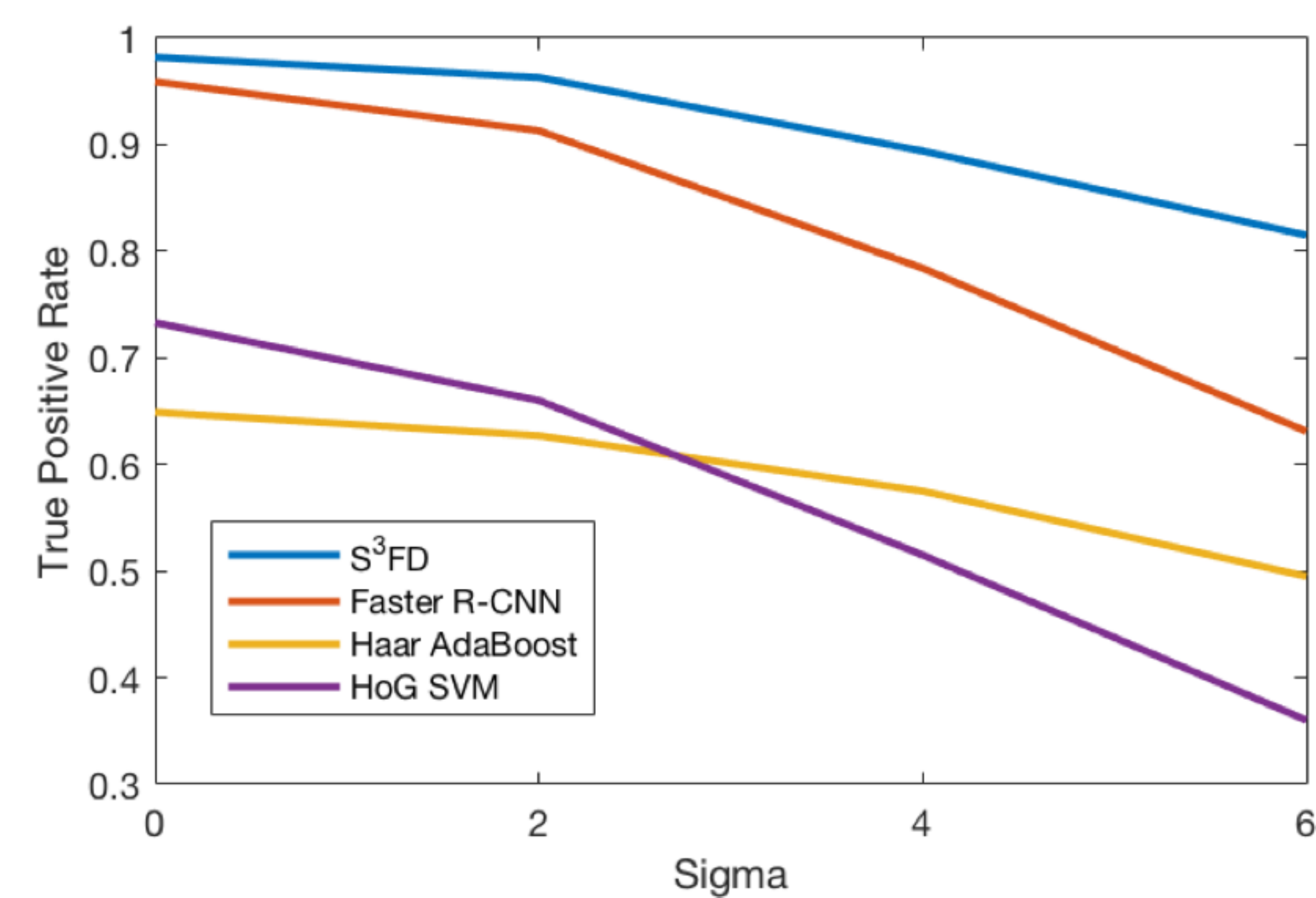}
        \caption{}
    \end{subfigure}%
    \hspace{5mm}
    \begin{subfigure}[t]{0.3\textwidth}
        \includegraphics[height=1.6in]{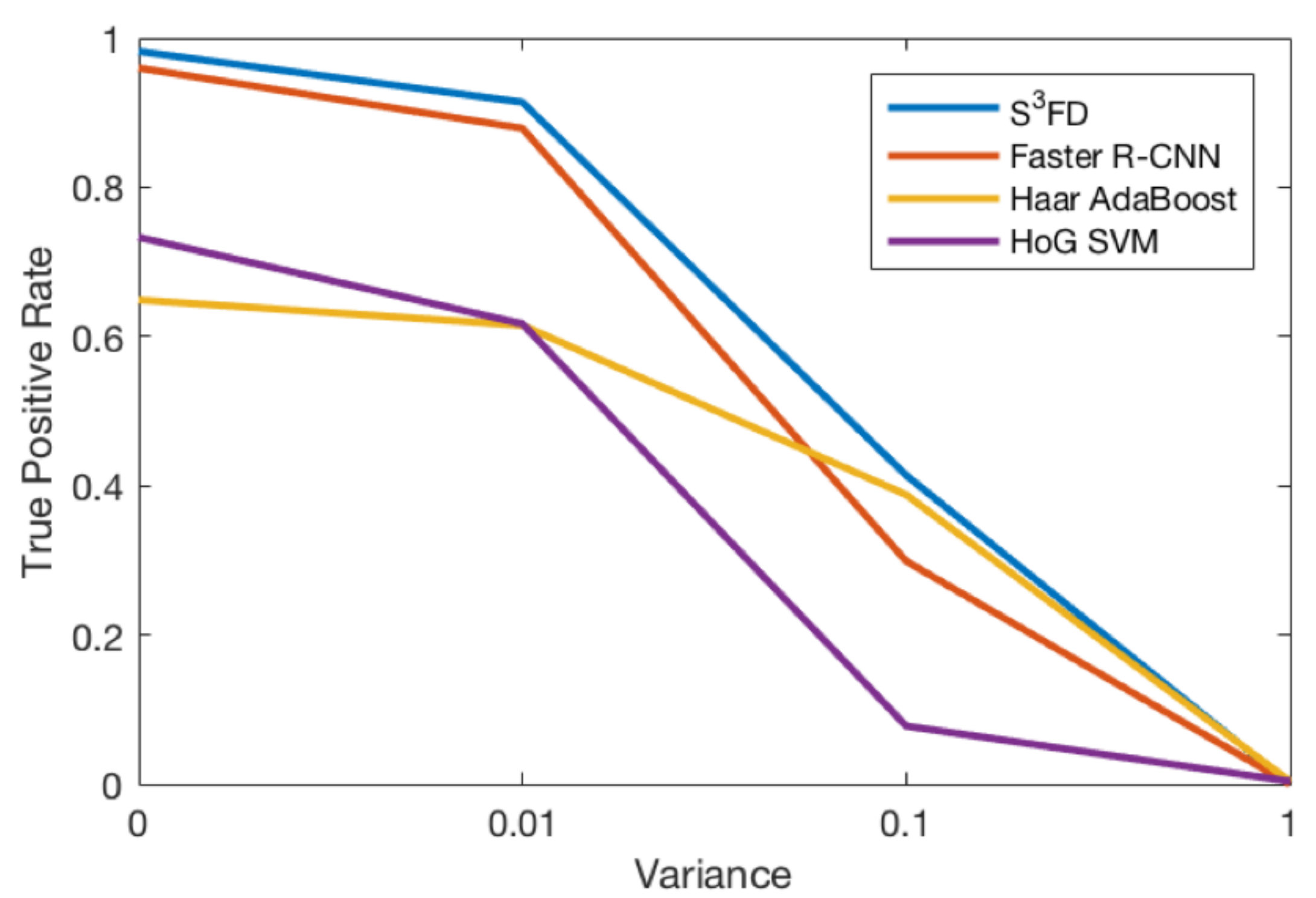}
        \caption{}
    \end{subfigure}
    \hspace{5mm}
    \begin{subfigure}[t]{0.3\textwidth}
        \includegraphics[height=1.6in]{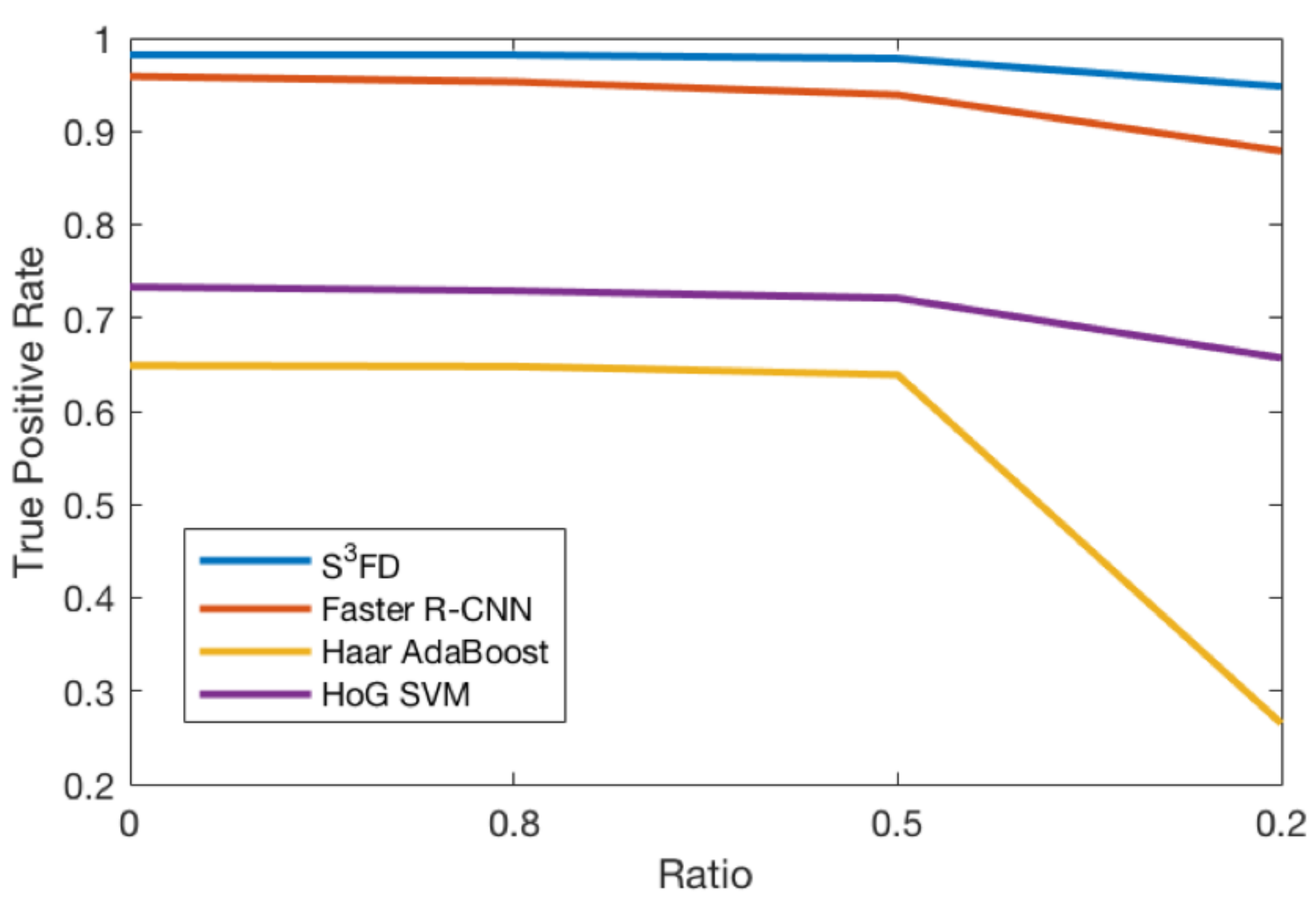}
        \caption{}
    \end{subfigure}
    \caption{Comparison of evaluation results  for all the four models tested. Performance degradation with (a) different levels of blur, (b) noise, and (c) decreasing brightness and contrast level.}
    \label{fig:all_comp}
    \vspace{-6mm}
\end{figure*}

\subsection{Models}
In this section, we introduce the face detectors we considered for evaluation. The first two models \cite{vj_face,dlib_face} exploit hand-crafted features. Viola-Jones detector \cite{vj_face} is a simple cascade model utilizing Haar features. It applied image pyramid with face templates of fixed size while testing. \cite{dlib_face} applies HoG features. Both of them are efficient for frontal face detection.

For deep learning models, we select faster R-CNN \cite{fasterrcnnface} and $S^3$FD \cite{sfd}. Faster R-CNN \cite{ren2015faster} introduces a region proposal network (RPN) to predict the positions of objects using anchor-based methods, and utilizes ROI pooling to extract features from the proposed regions. Since all the ROI with different sizes share the same classifier, it is a scale-invariant detector. The face detection model \cite{fasterrcnnface} based on faster R-CNN is transferred from a pretrained VGG16 \cite{vgg} on ImageNet \cite{krizhevsky2012imagenet}, and retrained on WIDER dataset \cite{yang2016wider}. 

$S^3$FD \cite{sfd} is an improved model of SSD \cite{liu2016ssd} with special designs for finding small faces. Compared to faster R-CNN, $S^3$FD and SSD utilize the features from multiple layers of deep networks for multi-scale detections. Mid-layers from lower-level to higher-level are associated with pre-defined anchors of doubling scales and stride sizes, and are connected with the corresponding prediction layers. Thus it is a scale-variant model. Like faster R-CNN, the backbone of $S^3$FD is also transferred from a pretrained VGG16 and further fine-tuned on WIDER Face. We select these two deep learning models because they represent scale-invariant and scale-variant detectors respectively, and are both transferred from a pre-trained VGG16, which is proved to be the most resilient to image distortions \cite{lq1}.

\begin{figure*}[t]
    \centering
        \centering
        \includegraphics[width=1\linewidth]{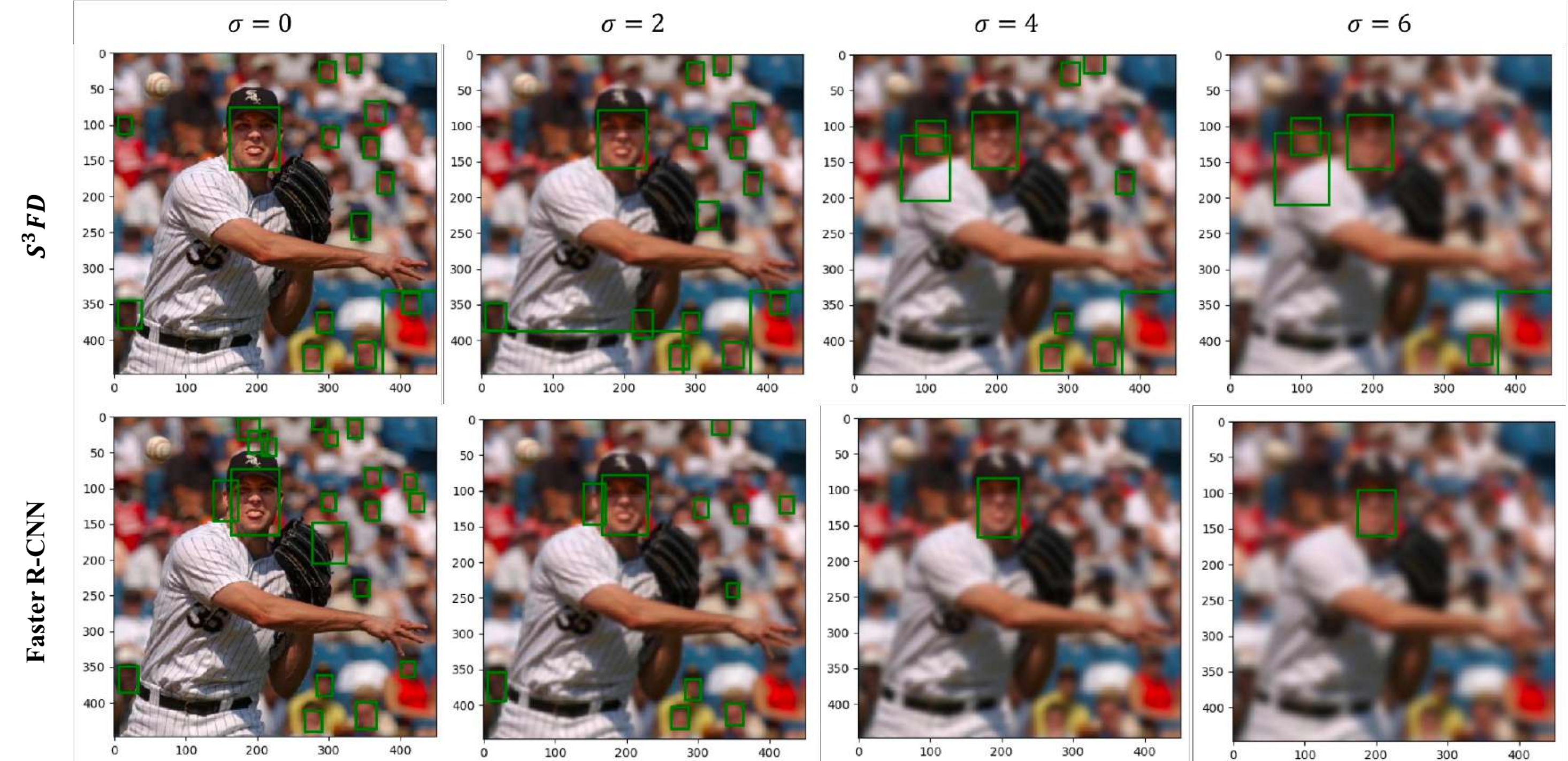}
    \caption{Detection results of $S^3$FD and faster R-CNN on various levels of blur. $S^3$FD achieves a better robustness for detecting blurry tiny faces because of utilizing more features from lower-level layers for detection. }
    \label{fig:detec}
    \vspace{-6mm}
\end{figure*}

\subsection{Dataset and Processing}
The dataset we utilize to evaluate is the benchmark FDDB \cite{fddb}. It contains 5171 faces in totally 2845 images. Each face is annotated by an ellipse bounding box. Since the output from most face detectors is rectangular box, we fit the ellipse using the rectangular boxes before evaluating the ROC curve. We apply the discrete  Receiver Operating Characteristic
(ROC) curve for comparison.

To acquire low-quality images, we process the original images in FDDB by three types of distortions. Some examples of the processed images are shown in Fig. \ref{fig:process}.

\subsubsection{Blur} Gaussian blur is applied to reduce the noise and high-frequency components of the images. Specifically, two-dimensional Gaussian functions with standard deviation 2, 4 and 6 are utilized to convolve with the images to form a Gaussian scale space. Subsampling is not applied to the processed images, thus we do not change the original resolution. Human is still capable of detecting larger faces from the images under severe blur.

\subsubsection{Noise} Gaussian white noise is added to the original FDDB images. The mean of the noise is zero, and the variance is set to 0.01, 0.1 and 1 respectively. With the highest noise level, it becomes harder for human to differentiate faces from the background pattern. 

\subsubsection{Brightness and Contrast} We limit the pixel values of the original images by shrinking the ranges. Specifically, we simultaneously decrease the brightness and contrast level by rescaling the pixel values with specific ratios 0.8, 0.5 and 0.2.

\section{Results and Discussions}

\subsection{Multi-scale Designs and Blur}
We first tested the four models on blurry images. Fig.~\ref{sfd} (a) shows ROC of $S^3$FD model evaluated on blurry images.  
For $S^3$FD, faster R-CNN, Haar Cascade and HoG, we report the true positive rate when the false positive samples are 2000, 750, 500, and 500 respectively. 
The comparison of each model while testing on images with different levels of blur is shown in Fig.~\ref{fig:all_comp} (a).

We found that both traditional and deep learning methods are not robust enough to blur testing samples, simply from the insufficient blurry features in the designed or learned filter banks. The multi-scale designs of face detection algorithms could not mitigate the negative influence of features, both for scale-invariant and scale-variant methods. Specifically, scale-invariant approaches like faster R-CNN applied the same detector for any scales, theoretically eliminated the influence of blur or feature resolution. However, faster R-CNN only extracted features of ROI from one single higher layer, which was influenced the most by a blurry input compared with lower layers. It makes detecting smaller blurry faces harder. Scale-variant detectors like $S^3$FD or SSD extracted features from multiple scale-specific layers including the lower layers, which are only slightly influenced by blur. According to Fig.~\ref{fig:all_comp}, we observe that $S^3$FD dropped more slowly than faster R-CNN because of utilizing more features from lower layers for detecting small faces. 

To further verify the above statement, we visualize some detection results as shown in Fig.~\ref{fig:detec}. The testing images contain a larger face on the foreground, and multiple blurry smaller faces on the background. We set the testing threshold of confidence to 0.1 for both of $S^3$FD and faster R-CNN to recall more possibilities. Both two models achieved a satisfactory detection performance for blurry faces, but as the overall image suffers more severe blur degradation, faster R-CNN failed to detect small faces when $\sigma = 4$, while $S^3$FD could still find some positive samples.   

\subsection{Noise and Contrast}
Fig. \ref{fig:all_comp} (b) shows the performance degradation when testing models on synthetic noisy images. We found that the detection efficiency of all the models are greatly influenced by additive noises, especially when the variance reaches 1, all the models could not detect any faces. However, for human, we could still possibly differentiate faces from background in the second row of Fig. \ref{fig:process}. We conjecture that images with or without noises contain greatly different visual cues for detection, which confused the pretrained network using noise-free features. Under this situation, the multi-scale designs of face detectors will not benefit the detections.

The results of evaluation on low-contrast and dark images is shown in Fig. \ref{fig:all_comp} (c). Different from the previous two situations, deep networks or traditional methods demonstrated better robustness because of the normalization process while testing. 

\section{Conclusions}
In this paper, we made a survey on face detection algorithms, and evaluated the representatives of them: Haar-like Adaboost cascade and HoG-SVM as traditional methods, and faster R-CNN and $S^3$FD as deep learning methods on low-quality images. We tested the performance degradation of the above models while changing the blur, noise or contrast level. The experiment results demonstrated that both hand-crafted and deeply learned features are quite sensitive to low-quality inputs. And compared to scale-invariant structure, scale-variant design of neural network extracting features from multiple layers could benefit the detection of blurry tiny faces. We hope our results will inspire more future work of quality-invariant face detectors for practical applications.

\section*{Acknowledgment}

This research work is supported in part by US Army Research Office grant W911NF-15-1-0317.


\bibliographystyle{ieeetr}
\bibliography{sample}

\end{document}